# Automatic Extraction of Bengali Root Verbs using Paninian Grammar

Arijit Das
Department of Computer Sc. And Engg.
Jadavpur University
Kolkata, India
arijit.mcse.ju@gmail.com

Tapas Halder
Department of Computer Sc. And Engg.
Jadavpur University
Kolkata, India
tapas.cs.ju@gmail.com

*Abstract*—**In this report, a supervised methodology has been presented to extract the root forms of the Bengali verbs using the grammatical rules proposed by Panini in Ashtadhyayi. The proposed system has been developed based on tense, person and morphological inflections of the verbs to find their root forms. The work has been executed in two phases: first, the surface level forms or inflected forms of the verbs have been classified into a certain numbers of groups of similar tense and person. For this task, a standard pattern, available in Bengali language has been used. Next, a set of rules have been applied to extract the root form from the surface level forms of a verb. The system has been tested on 10000 verbs collected from the Bengali text corpus developed in the TDIL project of the Govt. of India. The accuracy of the output has been achieved 99% which is verified by a linguistic expert. Root verb identification is a key step in semantic searching, multi-sentence search query processing, understanding the meaning of a language, disambiguation of word sense, classification of the sentences etc.**

## I. INTRODUCTION

In every language, the words in a sentence get morphologically inflected according to tense and person. The degree of inflection in English language is comparatively lower than the Asian languages. For example, the English word "play" has maximum four types of inflections: "play", "playing", "played" and "plays". But, in Asian languages like Bengali, Hindi, Tamil, Telugu, Marathi, Malayalam etc. the varieties of inflections are very large. For example, the Bengali verb " খেলা (khelā)" has so many inflected forms, as: খেলি (kheli), খেল (khela), খেলে (khele), খেলেন (khelen), খেলছে (khelchhe), খেলছিলে (khelchhile), খেলছিল (khelchila), খেলছিলেন (khelchhilen), খেলছিলাম (khelchhilam), খেলেছিল (khelechhila), খেলেছিলেন (khelechhilen), খেলব (khelba), খেলবে (khelbe), খেলবেন (khelben) and many more.

In every language, the verbs get inflected while those are used in the sentences following few specific rules related to person, tense and number. These rules were first proposed by Panini in Ashtadhyayi (for which lang). In this proposed work, those rules or sutras have been implemented to extract the root forms of the surface level Bengali verbs. Additionally, those surface level verbs have also been classified according to tense and person using supervised learning.

For case study, the proposed model has been applied on the surface level verbs in "Chalit" (colloquial) and "Sādhu" (chaste) form of Bengali language. The extraction of root verbs from the surface level verbs is the key step in every text processing task in computational linguistics.

## II. PROBLEM STATEMENT

Mathematically the problem can be stated in the following way:

Let, a root verb $\mu$ appears as $\alpha\mu\beta$ in its surface level form in different sentences, i.e.

$$\mu \rightarrow \alpha\mu\beta$$

The proposed model resolute $\mu$ from $\alpha\mu\beta$ after the classification w.r.t. its tense, person and number.

In case of Bengali language the morphological varieties of the words depend on geographical locations also. As, the Bengali word "খাওয়া" (khāoyā) appears in few localized forms, like- "খাবা" (khābā), "খাইবি" (khāibi), "খেয়ে নিবি" (kheye nibi), "খাইয়া ল" (khāiyā la) etc. The proposed algorithm can handle these types of occurrences of the verbs also.

## III. BRIEF SURVEY ON RELATED WORK

A lot of research works have been established till date in this domain. As this task is used as a subtask of stemming or lemmatization of a word, improvement in research works in this domain is still going on in different Indian languages.

The first stemmer was developed by Julie Beth Lovins [1] in 1968. Later the stemmer was improved by Martin Porter [2] in July, 1980 for English language.

A rule based Hindi lemmatizer has been developed by Snigdha Paul [3]. Paul developed an Inflectional Lemmatizer which generates the rules for extracting the suffixes and also added rules for generating a proper meaningful root word.

A Light Weight Stemmer for Bengali and its use in Spell Checker has been proposed by Md. Zahurul Islam [4]. Zahurul has presented a computationally inexpensive stemming algorithm for Bengali, which handles suffix removal in a domain independent way. The author used this algorithm to improve the performance of a spell checker and an information retrieval application.

Morphological Stemming Cluster Identification for Bangla has been developed by Amitava Das [5]. This proposed work describes the morphological parsing of Bangla (Bengali) words using clustering technique.

An approach to solve the semantic ambiguity problem for Bangla (Bengali) root words using Universal Networking Language (UNL) has been presented by M. F Mridha [6]. This work is more focused on solving the problems of Bangla (Bengali) sentences with semantic analysis, thereby improving the accuracy of the result.

Aloke Kumar Saha constructed an algorithm named Root Word (RW) Analysis Approach for Universal Words (UWs) in Bangla (Bengali) Language [7].

A lemmatization algorithm for Bengali has been developed and evaluated by Abhisek Chakrabarty and Utpal Garain [8]. The Bengali lemmatizer named as BenLem is found to be capable of handling both inflectional and derivational morphology in Bengali.

## IV. Proposed Approach

In the proposed approach, first the Bengali inflected verbs have been classified according to tense, based on suffix matching. The Suffixes (Vibhaktis) have been classified in ten major classes, like simple past, present continuous etc. Every class is labeled with a four digit bit pattern from a range of 0 (0000) to 9 (1001) (refer Table 1). The verbs have been classified by the system according to the suffixes and the longest matching code.

Next, the persons have been coded with 01, 10 and 11 for the first, second and third person respectively. In this step, the previously tense wise classified verbs are re-classified according to its person definitions.

After classifying the inflected verbs according to tense and person, a set of rules (sutras) from Paninian Logic (Summarized from Ashtadhyayi) are applied on these verbs. The rules are as follows:

- Number of characters
- Number of স্বরবর্ণ
- Number of ব্যঞ্জনবর্ণ
- Number of আ,ই,ঈ কার,র ফলা, ব ফলা etc.
- অনুসর্গ and উপসর্গ (prefix and suffix)
- ণ ত্ব বিধি
- ষ ত্ব বিধি
- Various rules for "*sandhi*"

  * Paninian rules depict: how the different forms of a root verb are changed according to various tenses and persons considering a set of well defined Sanskrit grammatical rules. As in the proposed approach, the Bengali verbs have been considered for the case study, a little modification has been applied on the basic rules to fit to the system.

## V. Text Normalization

The texts collected from the Bengali corpus are not adequately normalized. So, a manual text normalization procedure is applied before the work is carried out. The normalization steps include detachment of punctuation marks like single quote, tilde, double quote, parenthesis, comma, conversion of dissimilar fonts into similar one, removal of angular brackets, uneven spaces, broken lines, slashes, etc. from sentences, and identification of sentence terminal markers (i.e., "purnacched", note-of-exclamation, note-of-interrogation), etc.

## VI. Selection Of Inflected Verbs

After the text normalization, the normalized texts are passed through the Shallow parser (LTRC) to identify the inflected verbs.

## VII. Algorithm

The algorithm used for finding the root form of a verb is given below-

**Input**: List of inflected verbs.

**Output**: Root forms of the inflected verbs.

Step 1: Start

Step 2: All the inflected verbs are collected in a file.

Step 3: Repeat Step 4 for an individual verb.

Step 4: if any of the Suffixes (Vibhaktis) (refer Table 1) matches with the rear part of the verb,

Step 4.1: if any of the person identification rules match,

Step 4.1.1: Panini's rules are applied on those verbs to extract the root forms.

Step 5: Root forms are obtained as output.

Step 6: Stop.

## VIII. FLOWCHART

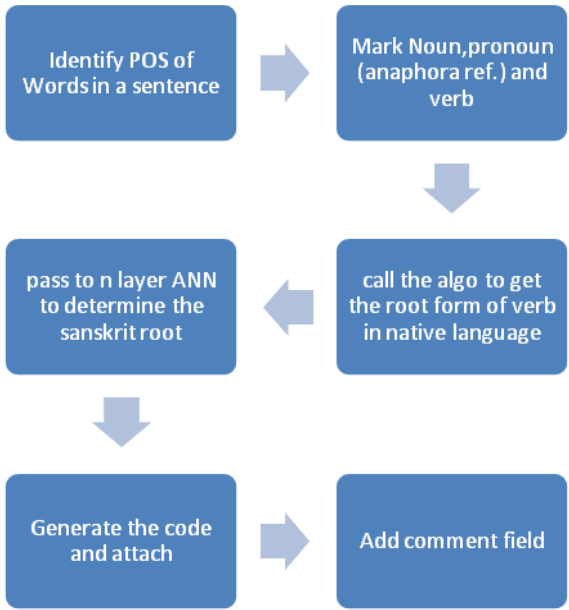

*Figure 1. Flowchart of the proposed approach.*

The list of suffixes used in this approach is given in Table 1.

Table 1. The list of suffixes used in this work.

| Tense | Type of tense | Suffices (বিভক্তি) |
|---|---|---|
| বর্তমান কাল (Present Tense) | সামান্য বর্তমান Simple Present | "ি","ো","েন","িস", "ই" |
| | ঘটমান বর্তমান Present Continuous | "ছি","িতেছি","ছে","িতেছে","ছ","িতেছ","ছেন" |
| | পুরাঘটিত বর্তমান Present Perfect | "েছি","িয়াছি","েছ","িয়াছ","েছে","েছেন" |
| অতীত কাল (Past Tense) | সামান্য অতীত Simple Past | "লাম","লুম","িলাম","িলুম","লে","িলে","লেন","িলেন" |
| | ঘটমান অতীত Past Continuous | "ছিলাম","ছিলুম","িতেছিলাম","িতেছিলুম" |
| | পুরাঘটিত অতীত Past Perfect | "েছিলাম","েছিলুম","িয়াছিলাম" |
| | নিত্যবৃত্ত অতীত Past Perfect Continuous | "তাম","তুম","িতাম","িতুম","তে" |
| ভবিষ্যৎ কাল (Future Tense) | সামান্য ভবিষ্যৎ Simple Future | "ব","িব","বে","িবে","বি","িবি" |
| | ঘটমান ভবিষ্যৎ Future Continuous | "তেথাকব","িতেথাকিব","তেথাকবে","িতেথাকিবে","িতেথাকিবি" |
| | পুরাঘটিত ভবিষ্যৎ Future Perfect | "েথাকব","িয়াথাকিব","েথাকবে" |

## IX. PERFORMANCE ANALYSIS

The algorithm has been tested on around 10,000 inflected verbs present in different forms in the corpus. The overall result is given below:

| Morphological class | Percentage of accuracy | Comment |
|---|---|---|
| Bangla cholit | 92% | urely modern official language used in official work in Bengal. |
| Bangla Sadhu | 88% | Old novels written in prose Bengali form. |
| Colloquial Bengali Language of Bangladesh | 85% | Generally known as Bangal language, taken as a whole but differs morphologically with geographic area. |
| Radh Bangla | 72% | Western part of westbengal part of Purulia, Bankura, Birbhum. |
| Mixed | 61% | Bengali Mixed with Hindi, English etc. |

## X. CLOSE OBSERVATIONS

It is observed that the algorithm has given an excellent result in case of Bengali Colloquial language (বাংলা চলিত ভাষা), like- "আমি কাল খেলা দেখতে যাব ।" and a good result in case of inflected Bangladeshi colloquial verbs, like- "আমি থাইতে যামু ।" and a satisfactory result in case of mixed forms of engali sentences, like-"আমি একটা bank account open করব ।" etc.

## XI. CONCLUSION AND FUTURE WORK

Though a considerable amount of work has been established in Bengali word stemming and lemmatization till date, most of them generally address all the parts-of-speeches. The verb lemmatization being the most difficult part in this domain pulls down the overall score of lemmatization. So, this methodology can be applied to improve the overall score of a Bengali lemmatizer.

This methodology has already been applied on huge collection of Bengali inflected verbs and perfect result has been achieved. This can also be applied on Indo Aryan languages.

As, in this methodology the tense and person of a verb are determined perfectly, it can be used for sentence and mood classification.

Root verb identification is a key step in semantic searching. It is a key step to understand the meaning of a language, disambiguate the meaning of a word and classify the sentences. The work has been developed for the modern and official form of Bengali language (colloquial form), "Sādhu Bhāshā" (chaste form), "Bānglā Bhāshā and "Rār Bonger Bhāshā". The other morphological forms as "অপভ্রংশ" (apabhransha) and old poetic forms are still required to be addressed.